\documentclass[10pt,twocolumn,letterpaper]{article}

\usepackage{cvpr} 
\usepackage{times}
\usepackage{epsfig}
\usepackage{graphicx}
\usepackage{amsmath}
\usepackage{amssymb}
\usepackage{tikz}

\usepackage{booktabs}

\usepackage[pagebackref=true,breaklinks=true,colorlinks,bookmarks=false]{hyperref}

\newcommand\copyrighttext{%
  \footnotesize \textcopyright 2025 IEEE. Personal use of this material is permitted.
  Permission from IEEE must be obtained for all other uses, in any current or future 
  media, including reprinting/republishing this material for advertising or promotional 
  purposes, creating new collective works, for resale or redistribution to servers or 
  lists, or reuse of any copyrighted component of this work in other works.
  }
\newcommand\copyrightnotice{%
\begin{tikzpicture}[remember picture,overlay]
\node[anchor=south,yshift=10pt] at (current page.south) {\fbox{\parbox{\dimexpr\textwidth-\fboxsep-\fboxrule\relax}{\copyrighttext}}};
\end{tikzpicture}%
}

\begin{document}

\title{Training-free Dimensionality Reduction via Feature Truncation:\\ Enhancing Efficiency in Privacy-preserving Multi-Biometric Systems}

\author{Florian Bayer, Maximilian Russo, Christian Rathgeb\\
da/sec–Biometrics and Internet-Security Research Group\\
Hochschule Darmstadt\\
{\tt\small florian.bayer@h-da.de}
}

\maketitle
\copyrightnotice
\thispagestyle{empty}

\begin{abstract}
  Biometric recognition is widely used, making the privacy and security of extracted templates a critical concern.
  Biometric Template Protection schemes, especially those utilizing Homomorphic Encryption, introduce significant computational challenges due to increased workload.
  Recent advances in deep neural networks have enabled state-of-the-art feature extraction for face, fingerprint, and iris modalities.
  The ubiquity and affordability of biometric sensors further facilitate multi-modal fusion, which can enhance security by combining features from different modalities.
  This work investigates the biometric performance of reduced multi-biometric template sizes.
  Experiments are conducted on an in-house virtual multi-biometric database, derived from DNN-extracted features for face, fingerprint, and iris, using the FRGC, MCYT, and CASIA databases.
  The evaluated approaches are (i) explainable and straightforward to implement under encryption, (ii) training-free, and (iii) capable of generalization.
  Dimensionality reduction of feature vectors leads to fewer operations in the Homomorphic Encryption (HE) domain, enabling more efficient encrypted processing while maintaining biometric accuracy and security at a level equivalent to or exceeding single-biometric recognition.
  Our results demonstrate that, by fusing feature vectors from multiple modalities, template size can be reduced by 67~\% with no loss in Equal Error Rate (EER) compared to the best-performing single modality.
\end{abstract}

\section{Introduction}

Alongside their widespread adoption, biometric recognition systems leveraging unique physiological and behavioral traits such as face, fingerprint, and iris have raised significant concerns regarding the privacy and security of stored features (templates) due to the irrevocable nature of biometric data. To address these issues, Biometric Template Protection (BTP) schemes have been developed, with Homomorphic Encryption (HE) emerging as a promising solution for secure, privacy-preserving biometric matching in the encrypted domain. However, the adoption of HE in practical biometric systems is hindered by its substantial computational overhead, particularly when processing high-dimensional feature vectors commonly produced by deep neural network (DNN) feature extractors. This challenge is further amplified in multi-biometric systems, where feature vectors from multiple modalities are fused to enhance recognition accuracy and security.

To mitigate this workload, dimensionality reduction techniques, such as feature vector truncation, can be employed to decrease template size and, consequently, the number of operations required in the encrypted domain. Importantly, simple and training-free truncation methods are desirable, as they offer transparency, ease of implementation, and generalizability across different biometric modalities and systems.

This work investigates the impact of simple feature vector truncation on the biometric performance and computational efficiency of multi-biometric systems protected by HE. By systematically reducing template dimensionality and evaluating the trade-off between accuracy and efficiency, we aim to demonstrate that secure and privacy-preserving multi-biometric recognition can be achieved without compromising performance. By leveraging the compatibility of DNN-based feature extractors across modalities, our method achieves high security through multi-modal fusion, and significantly reduces computational overhead in the encrypted domain without requiring additional training or complex transformations.

This work is organized as follows: \autoref{sec:related-work} briefly reviews related work, \autoref{sec:proposed-system} describes the proposed system and dimensionality reduction methods in detail, \autoref{sec:results} presents experimental results, and \autoref{sec:conclusion} concludes with a discussion and provides perspectives on future work.

\section{Related Work}\label{sec:related-work}

BTP has been a longstanding research focus due to the sensitive and irrevocable nature of biometric data~\cite{jainBiometricTemplateSecurity2008, barniPrivacyProtectionBiometricBased2015,Melzi-PET-CSUR-2024}. Early approaches include biometric cryptosystems~\cite{uludagBiometricCryptosystemsIssues2004} and cancelable biometrics~\cite{rathgebSurveyBiometricCryptosystems2011, patelCancelableBiometricsReview2015}, which aim to secure templates against inversion and cross-matching attacks. More recently, HE has emerged as a promising solution for privacy-preserving biometric matching, enabling computations directly on encrypted data~\cite{barniPrivacycompliantFingerprintRecognition2010, Boddeti-FaceBTP-FHE-BTAS-2018}. However, the computational overhead of HE remains a significant barrier to practical deployment, especially for high-dimensional deep feature representations.

Dimensionality reduction techniques have been widely studied to address efficiency and storage concerns in biometric systems. Principal Component Analysis (PCA)~\cite{turkEigenfacesRecognition1991}, Linear Discriminant Analysis (LDA)~\cite{belhumeurEigenfacesVsFisherfaces1997}, and more recently, deep learning-based methods~\cite{Singh20, OsorioRoig-StableHashFaceIdentification-TBIOM-2021, Talreja21} have been used to obtain compact and discriminative feature vectors. In the context of BTP, training-free methods such as truncation and quantization are attractive due to their performance, simplicity and generalizability. For instance,
Gentile \etal~\cite{gentileSLICShortlengthIris2009} introduced the Short-Length Iris Code (SLIC), showing that iris template dimensionality can be simply reduced without significant loss in recognition accuracy by means of feature correlation analysis.

Historically, feature-level fusion has been considered challenging due to the heterogeneous nature of feature representations and the use of different similarity measures across modalities~\cite{martiriFeatureFusionScheme2013}.
This is especially true for multi-biometric template protection systems, where alignment has been a fundamental challenge~\cite{rathgebMultiBiometricTemplateProtection2012}.
However, this situation has changed with the advent of DNN-based feature extractors, which provide compatible, fixed-length, and often normalized feature vectors for different biometric modalities.
This compatibility facilitates feature-level fusion and the application of dimensionality reduction techniques across modalities~\cite{jainScoreNormalizationMultimodal2005}.

Recent works have explored the trade-off between template size, recognition performance, and privacy protection in the context of secure multi-biometric systems~\cite{bauspiessImprovedHomomorphicallyEncrypted2022}. However, there remains a need for systematic studies on the impact of simple, explainable dimensionality reduction strategies on the efficiency and security of homomorphically encrypted multi-biometric recognition.

Our work builds on these foundations by evaluating the effect of training-free dimensionality reduction methods on the biometric accuracy and computational performance of multi-biometric systems protected by HE.

The amount of entropy contained in different biometric modalities varies considerably (depending on the sample quality and the employed feature extraction algorithm). For instance, face recognition systems are estimated to provide approximately 35 to 55 bits of entropy~\cite{adlerMeasureBiometricInformation2006}, while fingerprint recognition offers a slightly higher range of about 50 to 70 bits~\cite{youngEntropyFingerprints2013}.
Iris recognition, on the other hand, is known to provide significantly greater entropy, exceeding 200 bits~\cite{daugmanUnderstandingBiometricEntropy2023}.
For comparison, a 6-digit PIN contains only 20 bits of entropy, as it represents one out of one million possible combinations ($\log_2(1,000,000)$).

Multi-biometric systems typically assume statistical independence between modalities.
In other words, there is no information overlap between representations extracted from one modality (\eg, face) with another (\eg, iris).
This is desirable because combining independent sources contributes to overall entropy, thereby increasing resistance against attacks. For certain pairs of biometric modalities, \eg face and fingerprint, no correlations are to be expected. For other pairs of modalities, this desired independence is not that obvious. For instance, facial images contain the iris region, which enables matching face with iris images \cite{7026012}. However, modern face recognition systems (such as ArcFace~\cite{dengArcFaceAdditiveAngular2019}, MagFace~\cite{mengMagFaceUniversalRepresentation2021}) typically process facial images cropped to resolutions around 112$\times$112 pixels.
In effect, the iris region occupies only a few pixels and the spatial resolution does not contain meaningful iris texture.
Additionally, face recognition systems are trained on global properties rather than fine-grained texture.
In contrast, iris recognition systems (\eg~\cite{daugmanHowIrisRecognition2004,hafner_deep_2021}) use high-resolution close-up images of the eye region and discard surrounding information during segmentation. Consequently, the correlation between corresponding feature vectors is expected to be negligible. 
\section{Proposed System}\label{sec:proposed-system}
\begin{figure*}[ht]
  \centering
  \includegraphics[width=0.9\textwidth]{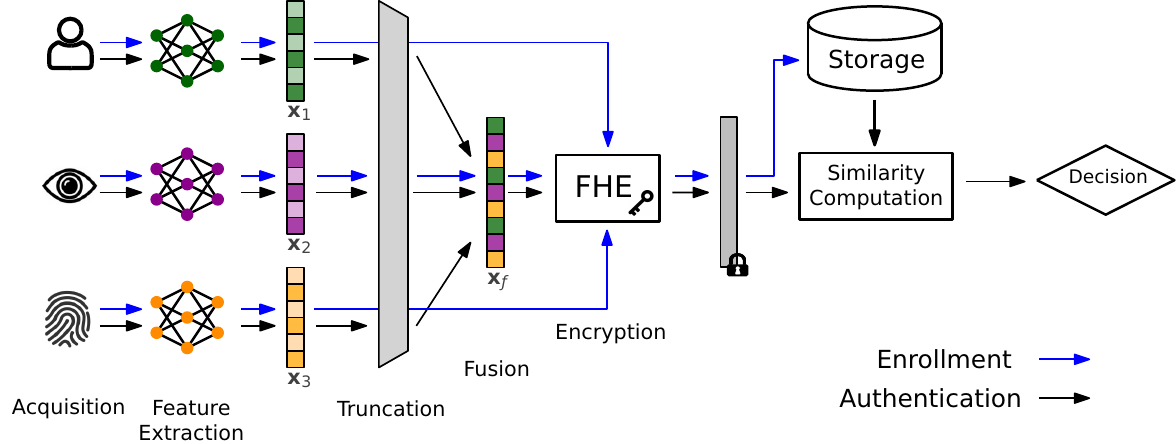}
  \caption{Overview of the proposed system.}
  \label{fig:proposed-system-overview}
\end{figure*}
The proposed system consists of three main components: (i) deep neural network (DNN)-based feature extraction, (ii) dimensionality reduction via  truncation strategies, and (iii) similarity computation performed under HE.
First, state-of-the-art DNNs are used to extract fixed-length feature vectors from biometric samples of different modalities.
Next, these feature vectors are reduced in size using simple, training-free truncation methods such as fractions, interleaving, or summation.
Finally, template matching is carried out in the encrypted domain using HE, ensuring privacy-preserving comparison without exposing sensitive biometric information.
This method allows storage of complete feature vectors during enrollment and enables partial matching, thereby increasing flexibility.
An overview of the system is depicted in \autoref{fig:proposed-system-overview}.
\subsection{Deep Feature Extraction and Comparison}
For the experimental evaluation, we selected three state-of-the-art DNN-based feature extractors, one for each modality (face, fingerprint, and iris), and analyzed the properties of their resulting embeddings.
All extractors produce 512-dimensional feature vectors, which are normalized such that their elements are centered around zero. This consistency in dimensionality and normalization facilitates straightforward fusion and comparison using a single metric across modalities, and enables the application of training-free dimensionality reduction techniques described in the following subsections.
Regarding biometric fusion, score-level fusion using the sum rule has been observed to be competitive~\cite{jainGuidelinesBestPractices2015}.

To ensure comparable performance across modalities, the quality of face images of the FRGCv2 database has been degraded by applying some noise and blurring to match the performance of iris and fingerprint recognition.
Therefore, all three modalities contribute approximately equally to recognition performance at the baseline dimension, even though the real entropy of characteristics differs~\cite{adlerMeasureBiometricInformation2006,youngEntropyFingerprints2013,daugmanUnderstandingBiometricEntropy2023}.

\subsection{Application to HE}
When HE-based BTP is applied, computational complexity for template comparison is increased.
A significant amount of time is required to perform homomorphic rotations \cite{Bayer25a}. 
Hence, by reducing the dimension of feature vectors, time can be saved.
All truncation strategies presented in this work were deliberately chosen for their compatibility with HE schemes. Specifically, these methods—truncation, interleaving, and sum-approaches—are characterized by their reliance on simple, index-based selection or addition operations, which can be efficiently implemented in the encrypted domain. Unlike more complex dimensionality reduction techniques that require matrix multiplications or data-dependent transformations, the proposed strategies incur minimal computational overhead when compared to the comparison process itself. This ensures that the efficiency gains achieved through dimensionality reduction are not offset by increased complexity in the homomorphic fusion implementation, thereby facilitating practical deployment in privacy-preserving biometric systems.
In \autoref{fig:proposed-system-overview}, the application of HE to truncated features is depicted.
In order to increase security, it is advisable to fuse templates before encryption~\cite{merkle_multi-modal_2012}.
Therefore, during enrollment, full templates are stored concatenated, and only truncated features are compared during authentication.
The DNN feature embeddings can be compared using the squared Euclidean distance (SED).
This distance can easily be evaluated by homomorphically computing the Hadamard product~\cite{Boddeti-FaceBTP-FHE-BTAS-2018}.
\subsection{Notation}
Throughout this work, we use the following mathematical notation: vectors are denoted in boldface, \eg, $\mathbf{v} \in \mathbb{R}^d$ represents a $d$-dimensional real-valued feature vector. For indexing, we use subscripts, \ie the $i$-th element of a vector $\mathbf{v}$ is denoted by $v_i$. Fractions or partitions of vectors are indicated by subscripts or parentheses, such as $\mathbf{v}_{\text{first}}$ for the first part, or $\mathbf{v}_{(i)}$ for the $i$-th fraction.
The term \emph{Truncation} was chosen to describe the process of reducing the feature vector length. While the mathematical concept of truncation typically refers to cutting off decimal places, and the Unix utility \textsf{truncate} refers to cutting off files after a specific length, the term is adapted here to signify a straightforward reduction in feature vector size. Unlike complex statistical dimensionality reduction techniques, truncation in this work is performed using an equal-width threshold for binarization. This approach is simple, generalizable, and training-free.
\subsection{Pre-processing}
\begin{figure}
  \centering
  \includegraphics[width=0.8\columnwidth]{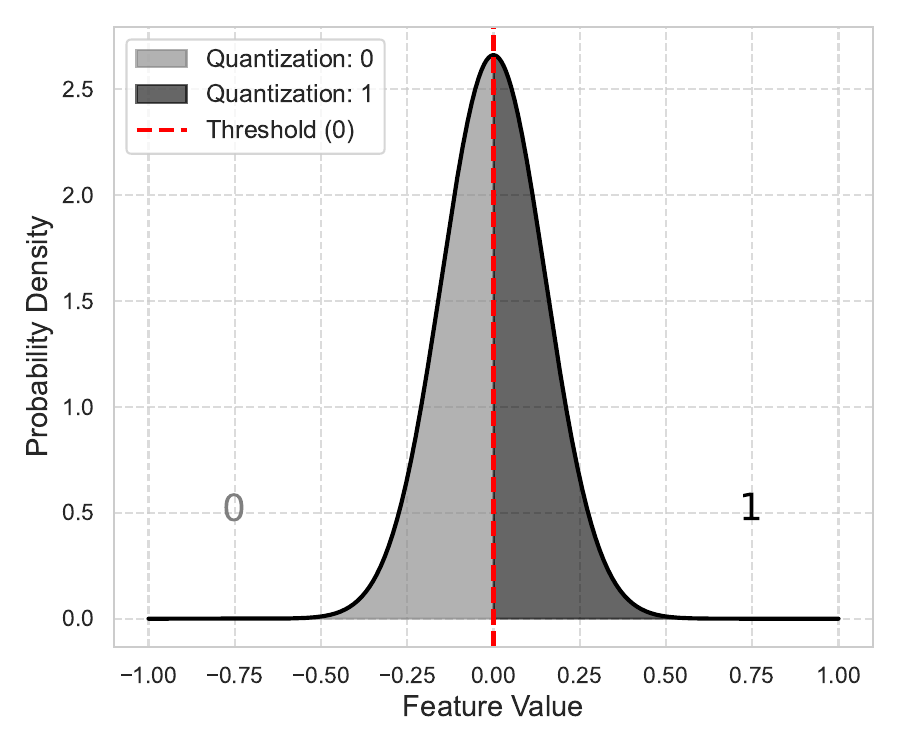}
  \caption{Example of quantization for threshold $0$.}
  \label{fig:quantization}
\end{figure}

In order to avoid expensive approximate arithmetic in the homomorphic domain, binarization is a viable approach. Different methods exist~\cite{drozdowskiBenchmarkingBinarisationSchemes2018}, but in the scope of this work we focus on a trivial single-bit encoding.
Mathematically speaking, we want to transform real-valued feature vector elements into binary:
\begin{equation}
  \operatorname{bin} : \mathbb{R} \to \{0, 1\}
\end{equation}
That is, given a feature vector element $x$ and a threshold $t$, output $0$ if the value is smaller than $t$ and $1$ otherwise, see \autoref{fig:quantization}:

\begin{equation}
  \operatorname{bin}(x) =
  \begin{cases}
    0 & \text{if } x < t,   \\
    1 & \text{if } x \text{ otherwise.}
  \end{cases}
\end{equation}

This reduces the amount of information needed to represent the feature vector at the cost of some accuracy. We can set this  threshold to $0$, which ideally is the same as the mean of the feature vector elements for a normalized output. This trivial approach called \emph{equal-width quantization} \cite{drozdowskiBenchmarkingBinarisationSchemes2018} may yield acceptable results if the distribution of feature element values follows a normal distribution. To make even better use of the available feature space, one can take the probability densities of the extracted features into account and allocate equal-probable portions but this requires training and thus we focus on the equal-width approach.
This approach can be generalized to more intervals:
Let $q=2^l$ denote the number of quantization intervals (\eg, $q=256$ for int8, $q=16$ for int4, $q=4$ for int2, $q=2$ for binary). The quantization function can be defined as:
\begin{equation}
  \operatorname{quant}_q(x) = \left\lfloor \frac{x - x_{\min}}{x_{\max} - x_{\min}} \cdot (q-1) \right\rfloor
\end{equation}
where $x_{\min}$ and $x_{\max}$ are the minimum and maximum values of the feature vector (or a fixed range, \eg, $[-1, 1]$ for normalized features).
\subsection{Truncation}
\paragraph{Fractions}
Let $\mathbf{v}^d_m$ be the $d$-dimensional feature vector of modality $m$. The vector is truncated to $1/x$ with $x \in \{2, 4\}$ representing halves and quarters, respectively.
%

%
%
%
To halve a vector, let $\mathbf{v} \in \mathbb{R}^d$ be a feature vector of length $d$.
We define the first half and second half of $\mathbf{v}$ as:
\begin{equation}
  \mathbf{v}_{\text{first}} = \left( v_1, v_2, \ldots, v_{d/2} \right)
\end{equation}
likewise, for the second vector:
\begin{equation}
  \mathbf{v}_{\text{second}} = \left( v_{d/2+1}, v_{d/2+2}, \ldots, v_d \right)
\end{equation}
where $d$ is assumed to be even.
In order to generalize to $k$-th fraction, for $k \in \mathbb{N}$, $k \mid d$ (\ie, $k$ divides $d$), the $i$-th fraction is:
\begin{equation}
  \mathbf{v}_{(i)} = \left( v_{(i-1)\frac{d}{k} + 1}, \ldots, v_{i\frac{d}{k}} \right), \quad i = 1, \ldots, k
\end{equation}
\paragraph{Interleaving}
For interleaving, we select every $2$nd, $4$th, and $8$th feature element from the original feature vector.
This can be formalized as generating indices using:
\begin{equation} \label{eq:interleaving_indices}
  i_k = \lfloor k \cdot \frac{d-1}{x-1} \rfloor, \quad k = 0, 1, \dots, x-1,
\end{equation}
where $d$ is the original feature vector size, $x \in \{2, 4, 8\}$ is the downsampling factor, and $i_k$ are the selected indices.
The resulting interleaved feature vector $\mathbf{v}_{[x]}$ is then:
\begin{equation} \label{eq:interleaved_vector}
  \mathbf{v}_{[x]} = \{ \mathbf{v}[i_k] \mid k = 0, 1, \dots, x-1 \}.
\end{equation}
For example, given $x = 4$, we select every $4$th element, resulting in a feature vector of size $d/4$.
\paragraph{Sums: Addition of Fractions and Modalities}
In addition to truncation and interleaving, we investigate a simple method for dimensionality reduction: the sum-approach. Instead of discarding parts of the feature vector, this approach combines information from different fractions (\eg, halves and quarters) by element-wise addition, resulting in a new, lower-dimensional representation.
Let $\mathbf{v} \in \mathbb{R}^d$ be a feature vector of length $d$, and let $k \mid d$ (i.e., $k$ divides $d$). We partition $\mathbf{v}$ into $k$ contiguous, non-overlapping fractions:
\begin{equation}
  \mathbf{v}_{(i)} = \left( v_{(i-1)\frac{d}{k} + 1}, \ldots, v_{i\frac{d}{k}} \right), \quad i = 1, \ldots, k
\end{equation}
The sum-approach constructs a new feature vector $\mathbf{v}_{\text{sum}}$ of length $d/k$ by summing the corresponding elements of each fraction:
\begin{equation}
  \mathbf{v}_{\text{sum}} = \sum_{i=1}^k \mathbf{v}_{(i)}
\end{equation}
where the sum is performed element-wise:
\begin{equation}
  \mathbf{v}_{\text{sum}}[j] = \sum_{i=1}^k v_{(i-1)\frac{d}{k} + j}, \quad j = 1, \ldots, \frac{d}{k}
\end{equation}
\par Example (Halves):
For $k=2$ (halves), the sum-approach yields:
\begin{equation}
  \mathbf{v}_{\text{sum}} = \mathbf{v}_{\text{first}} + \mathbf{v}_{\text{second}}
\end{equation}
with
\begin{equation}
  \mathbf{v}_{\text{sum}}[j] = v_j + v_{d/2 + j}, \quad j = 1, \ldots, d/2
\end{equation}
This approach preserves information from the entire feature vector while reducing the dimensionality and is straightforward to implement. It is particularly suitable for use with binarized vectors, as the sum can be interpreted as a count of "active" bits across fractions.
%

\begin{figure*}[ht]
  \centering
  \includegraphics[width=0.8\textwidth]{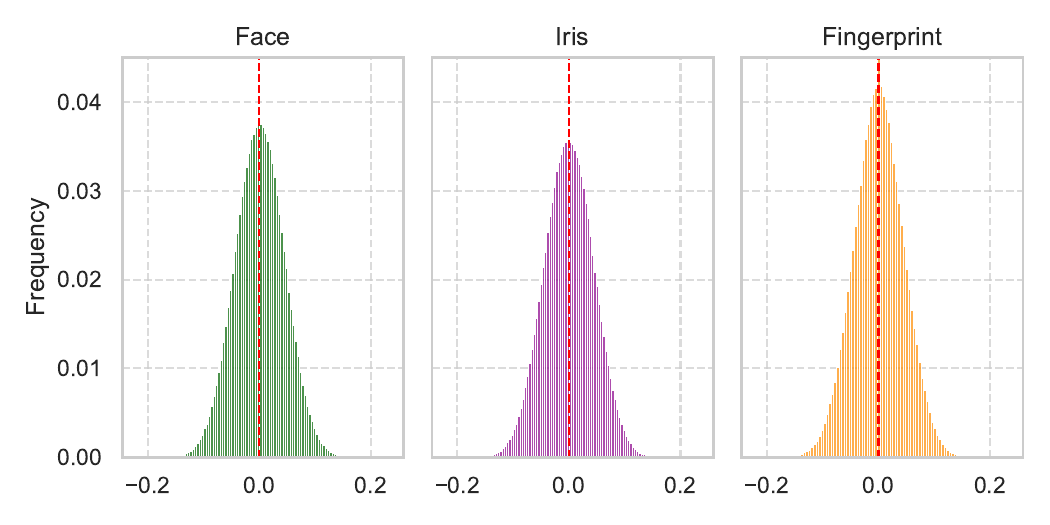}
  \caption{Distribution of feature vector elements for each modality (face, iris, fingerprint). The histograms show the value ranges and highlight the effect of normalization and binarization thresholds.}
  \label{fig:feature-vector-distributions}
\end{figure*}
\section{Results}\label{sec:results}
\subsection{Databases and Setup}
\autoref{tab:experiments} gives an overview of dimensionality reduction experiments conducted.
\begin{table}[h]
  \centering
  \caption{List of dimensionality reduction experiments.}
  \label{tab:experiments}
  \begin{tabular}{ll}
    \toprule
    \textbf{Experiment} & \textbf{Description}                       \\
    \midrule
    Binarization        & Quantization to $\{0,1\}$                  \\
    Fractions           & Truncation to $\frac{1}{n}$ of the vector, \\
    Interleaving        & Selecting every $n$-th element,            \\
    Sum-Approach        & Element-wise.                              \\
    \bottomrule
  \end{tabular}
\end{table}
\begin{figure}[ht]
  \centering
  \subfloat[Face]{%
    \includegraphics[height=2.1cm,clip,trim=0 40 0 40]{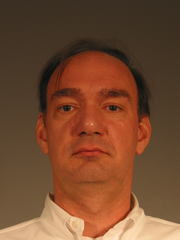}%
    \label{fig:face}}
  \hspace{0.01\columnwidth}
  \subfloat[Iris]{%
    \includegraphics[height=2.1cm,clip,trim=40 40 0 40]{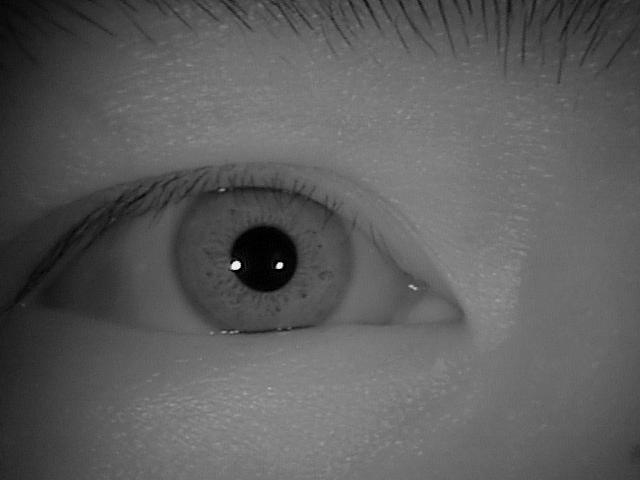}%
    \label{fig:iris}}
  \hspace{0.01\columnwidth}
  \subfloat[Fingerprint]{%
    \includegraphics[height=2.1cm,clip,trim=0 40 0 40]{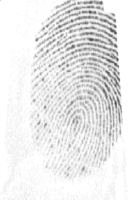}%
    \label{fig:fingerprint}}
  \caption{Sample images (first instance) for each biometric modality.}
  \label{fig:modalities}
\end{figure}
Sample images for each modality are shown in \autoref{fig:modalities}.
The virtual multi-biometric database was constructed by combining three established datasets, each representing a distinct modality: face, fingerprint, and iris.
Specifically, face images were sourced from the FRGCv2 database~\cite{phillipsOverviewFaceRecognition2005}, fingerprint samples (acquired using an optical sensor) from the MCYT database~\cite{ortega-garciaMCYTBaselineCorpus2003}, and iris images from the CASIA-Iris-Thousand dataset~\cite{casia}.
The feature extraction for each modality was performed using state-of-the-art deep neural network models: ArcFace~\cite{dengArcFaceAdditiveAngular2019} for face images, a re-implementation \cite{rohwedderBenchmarkingFixedLengthFingerprint2023} of DeepPrint~\cite{engelsmaLearningFixedLengthFingerprint2021} for fingerprint images, and the deep iris recognition model proposed in~\cite{hafner_deep_2021} for iris images. Each model outputs a 512-dimensional feature vector per sample, which serves as the basis for subsequent dimensionality reduction and fusion experiments.
The resulting multi-modal database comprises 533 virtual subjects with a total of 6,879 samples across all modalities.
When creating the database, left and right irises were distributed across virtual subjects, as well as different fingerprints, with each finger treated as a separate virtual subject.
For multi-biometric comparison, the number of comparisons was limited by the modality with the fewest available templates for each subject. This ensures that only samples present across all modalities are used in the evaluation.
In total, 1,759 mated comparisons and 141,778 non-mated comparisons were performed each of the experiments.
\subsection{Metrics}
In this work, we employ two standard metrics for evaluating biometric system performance: the Detection Error Tradeoff (DET) plot and the Equal Error Rate (EER).
The DET plot illustrates the trade-off between false match rate (FMR) and false non-match rate (FNMR) across varying decision thresholds.
The EER is a scalar summary metric derived from the DET curve. It is defined as the point at which FMR and FNMR are equal, representing a threshold where the system's false negative and false positive rates are balanced.
The EER follows a lower-is-better semantic, meaning that smaller EER values indicate improved biometric system performance.
Together, DET plots and EER provide both a detailed and an aggregate assessment of biometric recognition performance, enabling evaluation of the proposed methods. It is important to note that for the used SED-based comparison, the application of HE does not alter the recognition performance. This means that the reported performance rates are equal in unprotected and in the encrypted domain. 
\subsection{Feature Vector Distributions}
\autoref{fig:feature-vector-distributions} illustrates the distributions of the feature vector elements for each biometric modality. The histograms reveal that the feature values are approximately centered around zero, supporting the use of a threshold at zero for binarization. DNN-based feature extractors for all three produce real valued feature vectors normalized to $0$.
The distribution of fingerprint feature vector elements shows a concentration of values near zero, likely reflecting high similarity between instances due to the limited number of unique subjects in the dataset.
\subsection{Binarization}
For binarization, four different methods have been initially investigated: sign-based (threshold at $0$), min-max, and mean/median thresholding.
For all strategies except for min-max, the threshold was found to be $0$ due to the normalized distribution of feature values.
Min-max performed slightly better for face (1.92\% vs. 2.13\% EER) and fingerprint (1.37\% vs. 1.40\% EER), whereas sign performed better for iris (1.52\% vs. 1.47\% EER).
As the min-max method requires a priori knowledge of the extracted feature vectors, it was not considered further.
Therefore, we proceeded with the trivial, yet effective, sign-based approach.
\begin{figure}[ht]
  \centering
  \includegraphics[width=\columnwidth]{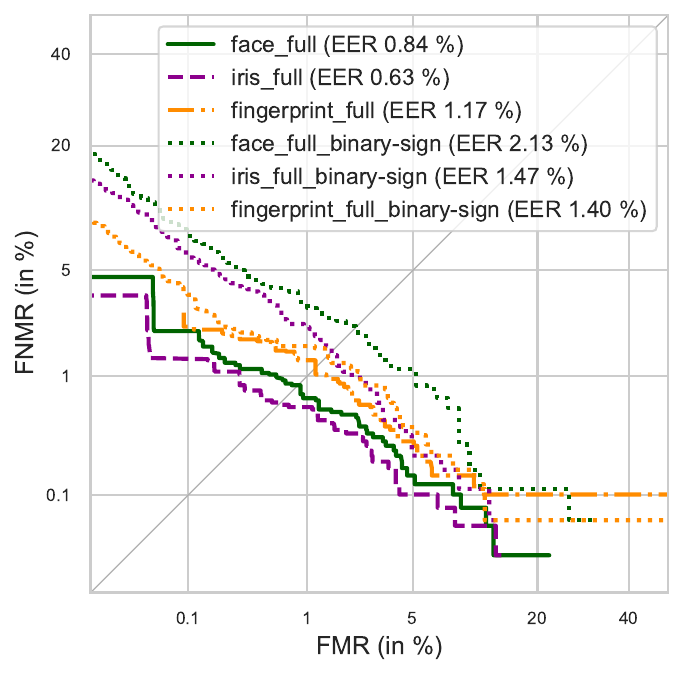}
  \caption{DET curve for baseline (full-length, real-valued and binary feature vectors) for all three modalities.}
  \label{fig:det-binary-float}
\end{figure}
\autoref{fig:det-binary-float} shows the DET curves for the baseline (real-valued) and binarized feature vectors, respectively. The baseline demonstrates the best achievable biometric performance, while binarization introduces a slight degradation in accuracy but enables efficient encrypted-domain processing.
\subsection{Fractions}
\begin{table}
	\centering
\caption{Results for fractions experiment (float) with mean ± std EER (\%) for each modality.}
\label{tab:experiments-fractions-float}
\resizebox{\linewidth}{!}{
\begin{tabular}{lcccc}
\toprule
\textbf{Dimension} & \textbf{All} & \textbf{Face} & \textbf{Finger} & \textbf{Iris} \\
\midrule
16 & 4.59 ± 5.35 & 18.67 ± 0.18 & 2.04 ± 0.15 & 8.31 ± 1.21 \\
32 & 1.98 ± 2.10 & 12.45 ± 0.56 & 1.58 ± 0.02 & 4.74 ± 0.13 \\
64 & 0.83 ± 0.24 & 7.49 ± 0.20 & 1.45 ± 0.05 & 2.83 ± 0.05 \\
128 & 0.31 ± 0.02 & 3.84 ± 0.48 & 1.37 ± 0.10 & 1.76 ± 0.11 \\
256 & 0.15 ± 0.12 & 2.00 ± 0.33 & 1.34 ± 0.11 & 1.33 ± 0.11 \\
512 (full) & 0.06  & 0.84  & 1.17  & 0.63  \\
\bottomrule
\end{tabular}
}
\end{table}

\begin{table}
	\centering
\caption{Results for fractions experiment (binary) with mean ± std EER (\%) for each modality.}
\label{tab:experiments-fractions-binary}
\resizebox{\linewidth}{!}{
\begin{tabular}{lcccc}
\toprule
\textbf{Dimension} & \textbf{All} & \textbf{Face} & \textbf{Iris} & \textbf{Finger} \\
\midrule
16 & 8.42 ± 10.88 & 25.88 ± 0.14 & 15.37 ± 0.66 & 6.49 ± 1.14 \\
32 & 4.38 ± 5.16 & 19.01 ± 0.68 & 9.25 ± 0.16 & 3.17 ± 0.41 \\
64 & 1.69 ± 0.80 & 12.02 ± 0.60 & 5.14 ± 0.06 & 2.14 ± 0.26 \\
128 & 0.60 ± 0.24 & 6.91 ± 0.34 & 3.13 ± 0.12 & 1.74 ± 0.20 \\
256 & 0.35 ± 0.08 & 3.46 ± 0.04 & 2.03 ± 0.00 & 1.54 ± 0.17 \\
512 (full) & 0.07  & 2.13  & 1.47  & 1.40  \\
\bottomrule
\end{tabular}
}
\end{table}

In \autoref{tab:experiments-fractions-float} and \autoref{tab:experiments-fractions-binary}, results for fractions are presented. In cases where the dimension is reduced, the standard deviation (std) is reported, depending on which parts of the feature vector is used. As expected, the biometric performance decreases with the dimension. Moreover, it is observable that for the fusion of the three modalities, the  dimension can be reduced to 128, while still outperforming the best single modality without applying dimensionality reduction. This holds for the use of the original as well as binarized feature vectors. 
\subsection{Interleaving}

In \autoref{tab:experiments-interleaving-float} and \autoref{tab:experiments-interleaving-binary}, results for the interleaving experiment are given. Here, similar trends as for the previous approach are observed.

\begin{figure}
  \includegraphics[width=\columnwidth]{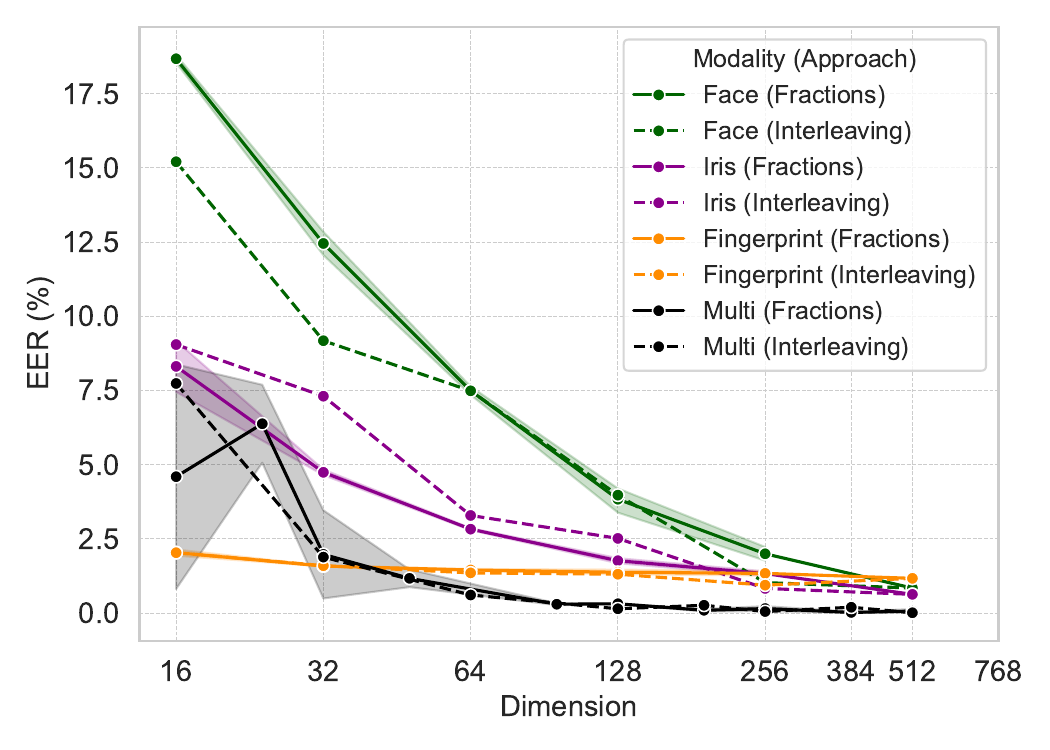}
  \caption{Feature Dimension vs. EER, fractions vs. interleaving (float).}
  \label{fig:float-fractions-vs-interleaving}
\end{figure}

\begin{figure}
  \includegraphics[width=\columnwidth]{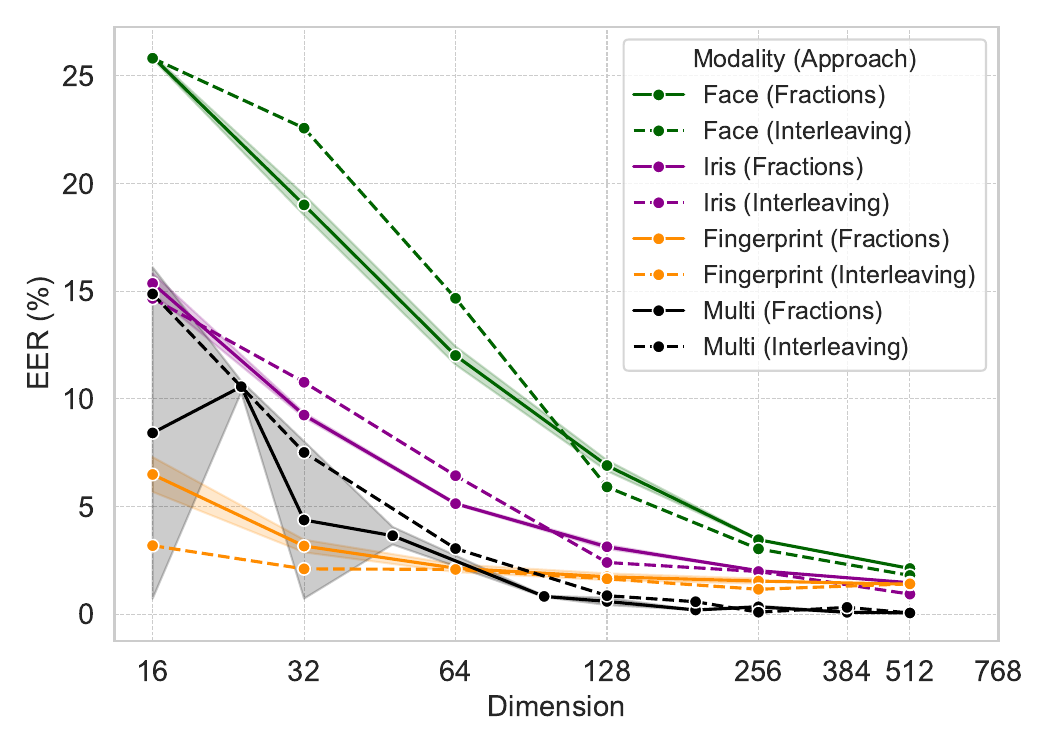}
  \caption{Feature Dimension vs. EER, fractions vs. interleaving (binary).}
  \label{fig:binary-fractions-vs-interleaving}
\end{figure}

\autoref{fig:float-fractions-vs-interleaving} and \autoref{fig:binary-fractions-vs-interleaving} show the comparison of EER versus feature dimension for both fractions and interleaving approaches using binarized feature vectors across all modalities and their fusion.
From the plots, it is evident that (i) the standard deviation for fractions is generally low for dimensions higher than $64$, indicating stable performance across different runs, and (ii) the fractions experiment achieves lower EER compared to interleaving for dimensions smaller than 128, probably due to loss in local context.
For fingerprint, EER increase is much smaller compared to other modalities, probably due to high similarity between instances from a single individual.
At dimensions greater than $96$, the multi-modal approach generally performs better than single modalities. For $d=256$ (half the length of an individual feature vector), the fused feature vectors achieved $\approx 10$ times lower EER than the best single modality.
\begin{table}
	\centering
\caption{Results for interleaving experiment (float) with EER (\%) for each modality.}
\label{tab:experiments-interleaving-float}
\begin{tabular}{lcccc}
\toprule
\textbf{Dimension} & \textbf{All} & \textbf{Face} & \textbf{Finger} & \textbf{Iris} \\
\midrule
16 & 7.73 & 15.20 & 2.03 & 9.05 \\
32 & 1.89 & 9.17 & 1.60 & 7.31 \\
64 & 0.61 & 7.49 & 1.35 & 3.29 \\
128 & 0.14 & 3.97 & 1.31 & 2.52 \\
256 & 0.05 & 1.03 & 0.94 & 0.83 \\
512 & 0.01 & 0.84 & 1.17 & 0.63 \\
\bottomrule
\end{tabular}
\end{table}

\begin{table}
	\centering
\caption{Results for interleaving experiment (binary) with EER (\%) for each modality.}
\label{tab:experiments-interleaving-binary}
\begin{tabular}{lcccc}
\toprule
\textbf{Dimension} & \textbf{All} & \textbf{Face} & \textbf{Finger} & \textbf{Iris} \\
\midrule
16 & 14.87 & 25.82 & 3.19 & 14.67 \\
32 & 7.52 & 22.58 & 2.11 & 10.78 \\
64 & 3.05 & 14.67 & 2.08 & 6.44 \\
128 & 0.87 & 5.91 & 1.65 & 2.41 \\
256 & 0.10 & 3.04 & 1.16 & 1.99 \\
512 & 0.06 & 1.81 & 1.41 & 0.94 \\
\bottomrule
\end{tabular}
\end{table}

\subsection{Sum}
The results of the sum-based strategy are listed in \autoref{tab:experiments-sum}. Addition of (fractions of) vectors has been observed to produce performance between the best and worst-performing fraction. For cases where it is unclear whether certain feature vector parts might contain more discriminative information, this approach is favorable. However, it should be noted that for this strategy it is advisable to apply the additions to the enrolled protected templates beforehand in order to accelerate the authentication process. 
\begin{table}
	\centering
	\caption{Results for sum experiment and full systems (float) with EER (\%).}
	\label{tab:experiments-sum}
	\begin{tabular}{lccccc}
		\toprule
		& \textbf{Multi} & \textbf{Sum} & \textbf{Face} & \textbf{Iris} & \textbf{Finger} \\
		\midrule
		Dimension & 1536 & 512 & 512 & 512 & 512 \\
		EER (\%)  & 0.03 & 0.09 & 0.84 & 0.63 & 1.17 \\
		\bottomrule
	\end{tabular}
\end{table}

\subsection{Summary of Results}
The experimental results demonstrate that simple dimensionality reduction techniques can significantly reduce multi-biometric template size without compromising recognition accuracy. Specifically, fusing features from face, fingerprint, and iris modalities allows the template dimension to be reduced by 67\% (from $1536$ to $512$) while maintaining accuracy in terms of EER compared to the best-performing single modality. Among the evaluated methods, both truncation (fractions) and interleaving approaches achieved low EERs, with multi-modal fusion consistently outperforming individual modalities. Binarization and quantization further reduced template size and computational workload, with only a minor impact on accuracy. These findings confirm that straightforward, training-free dimensionality reduction strategies are highly effective for privacy-preserving, efficient multi-biometric systems using HE.
It is important to note that the construction of the virtual multi-biometric database—where different fingers from the same subject are treated as separate virtual identities—likely leads to an overestimation of fingerprint recognition performance. Since multiple fingers from the same individual may share similar global characteristics or acquisition conditions, this setup can artificially inflate genuine match scores and reduce impostor variability, resulting in performance different from what would be observed in a real-world scenario with truly independent subjects.
%
%
Reducing from a 1536-dimensional real-valued vector to a 512-dimensional binary vector yields a $\sim$442$\times$ speedup in homomorphic comparison, attributed to fewer operations and efficient binary arithmetic. This is achieved without loss of accuracy: EER improves from 0.93\% (best single-modality, float) to 0.12\% (multi-modal, binary).
\section{Conclusion}\label{sec:conclusion}
In this work, we investigated the impact of simple and training-free dimensionality reduction techniques—namely truncation, interleaving, and sum-approaches on the biometric performance and computational efficiency of multi-biometric systems protected by HE.
Our experiments on a virtual multi-biometric database, constructed from DNN-extracted features for face, fingerprint, and iris modalities, demonstrate that significant template size reduction is achievable without loss in recognition accuracy.
Notably, by leveraging feature fusion, templates can be truncated to less than a third of their original size while still outperforming the best single-modality systems.
These findings highlight the potential of straightforward dimensionality reduction strategies to facilitate efficient and secure encrypted-domain biometric recognition, thereby improving privacy-preserving multi-biometric systems.
%
%
%

Several directions for future research emerge from this study:
\paragraph{Applicability to Other Template Protection Methods:} Although this work focuses on HE, the proposed dimensionality reduction methods—truncation, interleaving, and sum—are also applicable to other BTP schemes such as fuzzy vaults and cancelable biometrics. Reducing template size benefits efficiency and scalability across various privacy-preserving biometric systems, regardless of the underlying protection mechanism.
%
\paragraph{Datasets and Feature Extractors:} Extending the experiments to larger and more heterogeneous multi-biometric datasets, feature extractors and including additional modalities (e.g., voice, gait), would provide deeper insights into the generalizability and robustness of the proposed dimensionality reduction methods.
\paragraph{Exploring quantization intervals:} This work focused on binarization, but different quantization intervals are possible as well and their impact on biometric performance and computational efficiency should be investigated as well.
In addition to binarization, quantization to bit-widths higher than 1 is a promising approach for reducing both template size and computational workload while preserving more recognition performance than binarization whilst still allowing for more compact representations and efficient encrypted computation.
In the absence of sufficiently large real-world datasets, the use of synthetic data can serve as a valuable tool for preliminary experimentation and method development.
Synthetic data generation allows for controlled variation and augmentation, potentially increasing the diversity of training and evaluation scenarios.
However, synthetic data may not fully capture the complex statistical properties, noise characteristics, and inter-modal correlations present in real biometric data.
As a result, while synthetic data can help in scaling experiments and exploring edge cases, it may not be sufficient to conclusively demonstrate the robustness and generalizability of dimensionality reduction techniques as stable identity generation in synthetic data remains an open challenge for modalities other than face~\cite{boutrosIDiffFaceSyntheticbasedFace2023}.
Therefore, future research should prioritize the creation and use of large-scale, representative multi-biometric datasets, and carefully consider the limitations of synthetic data when interpreting experimental results.
\section*{Acknowledgments}
This research work has been funded by the German Federal Ministry of Education and Research and the Hessian Ministry of Higher Education, Research, Science and the Arts within their joint support of the National Research Center for Applied Cybersecurity ATHENE.
%
{\small
\bibliographystyle{ieee}
    \bibliography{egbib}

\begin{thebibliography}{10}\itemsep=-1pt

\bibitem{adlerMeasureBiometricInformation2006}
A.~Adler, R.~Youmaran, and S.~Loyka.
\newblock Towards a {{Measure}} of {{Biometric Information}}.
\newblock In {\em 2006 {{Canadian Conference}} on {{Electrical}} and {{Computer Engineering}}}, pages 210--213, May 2006.

\bibitem{barniPrivacycompliantFingerprintRecognition2010}
M.~Barni, T.~Bianchi, D.~Catalano, M.~Di~Raimondo, R.~D. Labati, P.~Failla, D.~Fiore, R.~Lazzeretti, V.~Piuri, A.~Piva, and F.~Scotti.
\newblock A privacy-compliant fingerprint recognition system based on homomorphic encryption and {{Fingercode}} templates.
\newblock In {\em 2010 {{Fourth IEEE International Conference}} on {{Biometrics}}: {{Theory}}, {{Applications}} and {{Systems}} ({{BTAS}})}, pages 1--7, Sept. 2010.

\bibitem{barniPrivacyProtectionBiometricBased2015}
M.~Barni, G.~Droandi, and R.~Lazzeretti.
\newblock Privacy {{Protection}} in {{Biometric-Based Recognition Systems}}: {{A}} marriage between cryptography and signal processing.
\newblock {\em IEEE Signal Processing Magazine}, 32(5):66--76, Sept. 2015.

\bibitem{bauspiessImprovedHomomorphicallyEncrypted2022}
P.~Bauspie{\ss}, J.~Olafsson, J.~Kolberg, P.~Drozdowski, C.~Rathgeb, and C.~Busch.
\newblock Improved {{Homomorphically Encrypted Biometric Identification Using Coefficient Packing}}.
\newblock In {\em 2022 {{International Workshop}} on {{Biometrics}} and {{Forensics}} ({{IWBF}})}, pages 1--6, Apr. 2022.

\bibitem{Bayer25a}
F.~Bayer and C.~Rathgeb.
\newblock {AMB-FHE}: Adaptive multi-biometric fusion with fully homomorphic encryption.
\newblock In {\em 2025 {{Int'l Workshop}} on {{Biometrics}} and {{Forensics}} ({{IWBF}})}, pages 1--6, Apr. 2025.

\bibitem{belhumeurEigenfacesVsFisherfaces1997}
P.~Belhumeur, J.~Hespanha, and D.~Kriegman.
\newblock Eigenfaces vs. {{Fisherfaces}}: Recognition using class specific linear projection.
\newblock {\em IEEE Transactions on Pattern Analysis and Machine Intelligence}, 19(7):711--720, July 1997.

\bibitem{Boddeti-FaceBTP-FHE-BTAS-2018}
V.~N. Boddeti.
\newblock Secure face matching using fully homomorphic encryption.
\newblock In {\em Proc. {{Intl}}. {{Conf}}. on Biometrics Theory, Applications and Systems ({{BTAS}})}, pages 1--10. IEEE, 2018.

\bibitem{boutrosIDiffFaceSyntheticbasedFace2023}
F.~Boutros, J.~H. Grebe, A.~Kuijper, and N.~Damer.
\newblock {{IDiff-Face}}: {{Synthetic-based Face Recognition}} through {{Fizzy Identity-Conditioned Diffusion Models}}.
\newblock In {\em 2023 {{IEEE}}/{{CVF International Conference}} on {{Computer Vision}} ({{ICCV}})}, pages 19593--19604, Oct. 2023.

\bibitem{casia}
{Chinese Academy of Sciences Institute of Automation}.
\newblock {{CASIA Iris Image Database}}, 2004.

\bibitem{daugmanHowIrisRecognition2004}
J.~Daugman.
\newblock How iris recognition works.
\newblock {\em IEEE Transactions on Circuits and Systems for Video Technology}, 14(1):21--30, Jan. 2004.

\bibitem{daugmanUnderstandingBiometricEntropy2023}
J.~Daugman.
\newblock Understanding {{Biometric Entropy}} and {{Iris Capacity}}: {{Avoiding Identity Collisions}} on {{National Scales}}, Aug. 2023.

\bibitem{dengArcFaceAdditiveAngular2019}
J.~Deng, J.~Guo, N.~Xue, and S.~Zafeiriou.
\newblock {{ArcFace}}: {{Additive Angular Margin Loss}} for {{Deep Face Recognition}}.
\newblock In {\em 2019 {{IEEE}}/{{CVF Conference}} on {{Computer Vision}} and {{Pattern Recognition}} ({{CVPR}})}, pages 4685--4694, June 2019.

\bibitem{drozdowskiBenchmarkingBinarisationSchemes2018}
P.~Drozdowski, F.~Struck, C.~Rathgeb, and C.~Busch.
\newblock Benchmarking {{Binarisation Schemes}} for {{Deep Face Templates}}.
\newblock In {\em 2018 25th {{IEEE International Conference}} on {{Image Processing}} ({{ICIP}})}, pages 191--195, Oct. 2018.

\bibitem{engelsmaLearningFixedLengthFingerprint2021}
J.~J. Engelsma, K.~Cao, and A.~K. Jain.
\newblock Learning a {{Fixed-Length Fingerprint Representation}}.
\newblock {\em IEEE Transactions on Pattern Analysis and Machine Intelligence}, 43(6):1981--1997, June 2021.

\bibitem{gentileSLICShortlengthIris2009}
J.~E. Gentile, N.~Ratha, and J.~Connell.
\newblock {{SLIC}}: {{Short-length}} iris codes.
\newblock In {\em 2009 {{IEEE}} 3rd {{International Conference}} on {{Biometrics}}: {{Theory}}, {{Applications}}, and {{Systems}}}, pages 1--5, Sept. 2009.

\bibitem{hafner_deep_2021}
A.~Hafner, P.~Peer, {\v Z}.~Emer{\v s}i{\v c}, and M.~Vitek.
\newblock Deep {{Iris Feature Extraction}}.
\newblock In {\em 2021 {{International Conference}} on {{Artificial Intelligence}} in {{Information}} and {{Communication}} ({{ICAIIC}})}, pages 258--262, Apr. 2021.

\bibitem{jainGuidelinesBestPractices2015}
A.~Jain, B.~Klare, and A.~Ross.
\newblock Guidelines for best practices in biometrics research.
\newblock In {\em 2015 {{International Conference}} on {{Biometrics}} ({{ICB}})}, pages 541--545, Phuket, Thailand, May 2015. IEEE.

\bibitem{jainScoreNormalizationMultimodal2005}
A.~Jain, K.~Nandakumar, and A.~Ross.
\newblock Score normalization in multimodal biometric systems.
\newblock {\em Pattern Recognition}, 38(12):2270--2285, Dec. 2005.

\bibitem{jainBiometricTemplateSecurity2008}
A.~K. Jain, K.~Nandakumar, and A.~Nagar.
\newblock Biometric {{Template Security}}.
\newblock {\em EURASIP Journal on Advances in Signal Processing}, 2008(1):579416, 2008.

\bibitem{7026012}
R.~Jillela and A.~Ross.
\newblock Matching face against iris images using periocular information.
\newblock In {\em Int'l Conf. on Image Processing (ICIP)}, pages 4997--5001, 2014.

\bibitem{martiriFeatureFusionScheme2013}
E.~Martiri, K.~Sevrani, and B.~Yang.
\newblock A feature fusion scheme for multi-biometric template protec.
\newblock {\em IFAC Proceedings Volumes}, 46(8):170--175, Jan. 2013.

\bibitem{Melzi-PET-CSUR-2024}
P.~Melzi, C.~Rathgeb, R.~Tolosana, R.~Vera, and C.~Busch.
\newblock An overview of privacy-enhancing technologies in biometric recognition.
\newblock {\em {ACM} Computing Surveys ({CSUR})}, May 2024.

\bibitem{mengMagFaceUniversalRepresentation2021}
Q.~Meng, S.~Zhao, Z.~Huang, and F.~Zhou.
\newblock {{MagFace}}: {{A Universal Representation}} for {{Face Recognition}} and {{Quality Assessment}}.
\newblock In {\em 2021 {{IEEE}}/{{CVF Conference}} on {{Computer Vision}} and {{Pattern Recognition}} ({{CVPR}})}, pages 14220--14229, Nashville, TN, USA, June 2021. IEEE.

\bibitem{merkle_multi-modal_2012}
J.~Merkle, T.~Kevenaar, and U.~Korte.
\newblock Multi-modal and multi-instance fusion for biometric cryptosystems.
\newblock In {\em 2012 {{BIOSIG}} - {{Proceedings}} of the {{International Conference}} of {{Biometrics Special Interest Group}} ({{BIOSIG}})}, pages 1--6, Sept. 2012.

\bibitem{ortega-garciaMCYTBaselineCorpus2003}
J.~{Ortega-Garcia}, J.~{Fierrez-Aguilar}, D.~Simon, J.~Gonzalez, M.~{Faundez-Zanuy}, V.~Espinosa, A.~Satue, I.~Hernaez, J.-J. Igarza, C.~Vivaracho, D.~Escudero, and Q.-I. Moro.
\newblock {{MCYT}} baseline corpus: A bimodal biometric database.
\newblock {\em IEE Proceedings - Vision, Image, and Signal Processing}, 150(6):395, 2003.

\bibitem{OsorioRoig-StableHashFaceIdentification-TBIOM-2021}
D.~Osorio-Roig, C.~Rathgeb, P.~Drozdowski, and C.~Busch.
\newblock Stable hash generation for efficient privacy-preserving face identification.
\newblock {\em IEEE Trans. on Biometrics, Behavior, and Identity Science ({TBIOM})}, 2021.

\bibitem{patelCancelableBiometricsReview2015}
V.~M. Patel, N.~K. Ratha, and R.~Chellappa.
\newblock Cancelable {{Biometrics}}: {{A}} review.
\newblock {\em IEEE Signal Processing Magazine}, 32(5):54--65, Sept. 2015.

\bibitem{phillipsOverviewFaceRecognition2005}
P.~Phillips, P.~Flynn, T.~Scruggs, K.~Bowyer, J.~Chang, K.~Hoffman, J.~Marques, J.~Min, and W.~Worek.
\newblock Overview of the face recognition grand challenge.
\newblock In {\em 2005 {{IEEE Computer Society Conference}} on {{Computer Vision}} and {{Pattern Recognition}} ({{CVPR}}'05)}, volume~1, pages 947--954 vol. 1, June 2005.

\bibitem{rathgebMultiBiometricTemplateProtection2012}
C.~Rathgeb and C.~Busch.
\newblock Multi-{{Biometric Template Protection}}: {{Issues}} and {{Challenges}}.
\newblock In J.~Yang, editor, {\em New {{Trends}} and {{Developments}} in {{Biometrics}}}. InTech, Nov. 2012.

\bibitem{rathgebSurveyBiometricCryptosystems2011}
C.~Rathgeb and A.~Uhl.
\newblock A survey on biometric cryptosystems and cancelable biometrics.
\newblock {\em EURASIP Journal on Information Security}, 2011(1):3, Dec. 2011.

\bibitem{rohwedderBenchmarkingFixedLengthFingerprint2023}
T.~Rohwedder, D.~{Osorio-Roig}, C.~Rathgeb, and C.~Busch.
\newblock Benchmarking {{Fixed-Length Fingerprint Representations Across Different Embedding Sizes}} and {{Sensor Types}}.
\newblock In {\em 2023 {{International Conference}} of the {{Biometrics Special Interest Group}} ({{BIOSIG}})}, pages 1--6, Sept. 2023.

\bibitem{Singh20}
A.~Singh, P.~Gaurav, C.~Vashist, A.~Nigam, and R.~P. Yadav.
\newblock Ihashnet: Iris hashing network based on efficient multi-index hashing.
\newblock In {\em International Joint Conference on Biometrics (IJCB)}, pages 1--9, 2020.

\bibitem{Talreja21}
V.~Talreja, M.~C. Valenti, and N.~M. Nasrabadi.
\newblock Deep hashing for secure multimodal biometrics.
\newblock {\em IEEE Transactions on Information Forensics and Security}, 16:1306--1321, 2021.

\bibitem{turkEigenfacesRecognition1991}
M.~Turk and A.~Pentland.
\newblock Eigenfaces for {{Recognition}}.
\newblock {\em Journal of Cognitive Neuroscience}, 3(1):71--86, Jan. 1991.

\bibitem{uludagBiometricCryptosystemsIssues2004}
U.~Uludag, S.~Pankanti, S.~Prabhakar, and A.~Jain.
\newblock Biometric cryptosystems: Issues and challenges.
\newblock {\em Proceedings of the IEEE}, 92(6):948--960, June 2004.

\bibitem{youngEntropyFingerprints2013}
M.~R. Young, S.~J. Elliott, C.~J. Tilton, and J.~E. Goldman.
\newblock Entropy of {{Fingerprints}}.
\newblock {\em International Journal of Computer Science Engineering and Technology}, 3(2):43--47, 2013.

\end{thebibliography}
  }

\end{document}